\documentclass[10pt, conference, letterpaper]{IEEEtran}
\IEEEoverridecommandlockouts
\usepackage{cite}
\usepackage{amsmath,amssymb,amsfonts}
\usepackage{algorithm}  
\usepackage{algorithmicx}  
\usepackage{algpseudocode}  
\usepackage{graphicx}
\usepackage{textcomp}
\usepackage{xcolor}
\usepackage{verbatim}
\usepackage{amsthm}
\theoremstyle{definition}

\usepackage{multirow}
\usepackage{listings}
\usepackage{caption}
\usepackage{xcolor}
\usepackage{subfigure}
	
\definecolor{codegreen}{rgb}{0,0.6,0}
\definecolor{codegray}{rgb}{0.5,0.5,0.5}
\definecolor{codepurple}{rgb}{0.58,0,0.82}
\definecolor{backcolour}{rgb}{0.95,0.95,0.92}

\lstdefinestyle{mystyle}{
	frame=top,frame=bottom,
    backgroundcolor=\color{backcolour},   
    commentstyle=\color{codegreen},
    keywordstyle=\color{magenta},
    numberstyle=\tiny\color{codegray},
    stringstyle=\color{codepurple},
    basicstyle=\ttfamily\footnotesize,
    breakatwhitespace=false,         
    breaklines=true,                 
    captionpos=t,                    
    keepspaces=true,                 
    numbers=left,                    
    numbersep=5pt,                  
    showspaces=false,
    showtabs=true,                
    showstringspaces=false,
    showtabs=false,                  
    tabsize=2
}
\newtheorem{theorem}{Theorem}[section]

\newtheorem{lemma}[theorem]{Lemma}
\newtheorem{assumption}{Assumption}[section]

\def\BibTeX{{\rm B\kern-.05em{\sc i\kern-.025em b}\kern-.08em
    T\kern-.1667em\lower.7ex\hbox{E}\kern-.125emX}}
\begin{document}

\title{\LARGE DS-Sync: Addressing Network Bottlenecks with Divide-and-Shuffle Synchronization for Distributed DNN Training\\}

\author{Weiyan Wang$^1$, Cengguang Zhang$^1$, Liu Yang$^1$, Kai Chen$^1$, Kun Tan$^2$\\
$^1$iSING Lab, Hong Kong University of Science and Technology, $^2$Huawei
}

\maketitle

\begin{abstract}
Bulk synchronous parallel (BSP) is the de-facto paradigm for distributed DNN training in today's production clusters. However, due to the global synchronization nature, its performance can be significantly influenced by network bottlenecks caused by either static topology heterogeneity or dynamic bandwidth contentions. Existing solutions, either system-level optimizations strengthening BSP (e.g., Ring or Hierarchical All-reduce) or algorithmic optimizations replacing BSP (e.g., ASP or SSP, which relax the global barriers), do not completely solve the problem, as they may still suffer from communication inefficiency or risk convergence inaccuracy. 

In this paper, we present a novel divide-and-shuffle synchronization (DS-Sync) to realize communication efficiency without sacrificing convergence accuracy for distributed DNN training. At its heart, by taking into account the network bottlenecks, DS-Sync improves communication efficiency by dividing workers into non-overlap groups to synchronize independently in a bottleneck-free manner. Meanwhile, it maintains convergence accuracy by iteratively shuffling workers among different groups to ensure a global consensus. We theoretically prove that DS-Sync converges properly in non-convex and smooth conditions like DNN. We further implement DS-Sync and integrate it with PyTorch, and our testbed experiments show that DS-Sync can achieve up to $94\%$ improvements on the end-to-end training time with existing solutions while maintaining the same accuracy.

\end{abstract}


\section{Introduction}

\begin{figure*}[ht]
  	\centering

  	\includegraphics[width=0.92\linewidth]{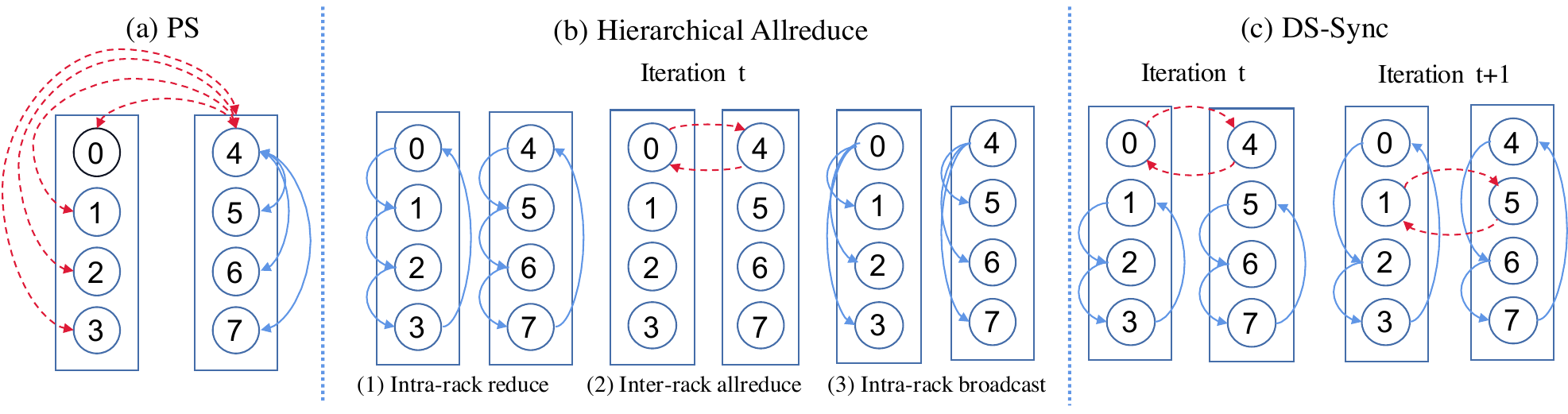}
  	\caption{Comparison in the setting of two racks with upper-level links indicated by red dashed lines. (a) In PS(widely used in BSP, ASP, and SSP), some workers are in different racks from the PS in worker 4, thus suffering from inter-rack bottlenecks when pushing to or pulling from the PS. (b) Topology-aware Hierarchical Allreduce decomposes the global communication into three serial steps, including intra-rack reduce, inter-rack allreduce, and intra-rack broadcast. While it mitigates traffic over the inter-rack bottleneck, it is still a global synchronization process of three serial communication steps. (c) DS-Sync divides workers into different sized groups to synchronize independently and shuffles group members every iteration. It fully utilizes intra-rack bandwidth while reducing the traffic and communication latency over the bottleneck. More importantly, all the groups synchronize {\em in parallel} as opposed to {\em sequentially} as the Hierarchical Allreduce.}
  \label{fig:ds_sync}
  \end{figure*}

In today's production training clusters, BSP is the {\em de-facto} paradigm to train DNN models with large datasets across different workers~\cite{DBLP:conf/usenix/JeonVPQXY19,DBLP:conf/osdi/ZhaoHYZYZYLWXW20,DBLP:conf/mlsys/LuoWKCN20,DBLP:conf/osdi/XiaoBRSKHPPZZYZ18}. In every iteration, BSP enforces a global synchronization to aggregate gradients from all workers and then distribute them back to each worker so that it updates model parameters as exactly the same way of training on the single machine~\cite{osdi:parameterserver}. However, there are network bottlenecks caused by either physical network oversubscription or dynamic bandwidth contentions in real-world environments. The synchronization process for some workers may be significantly affected (as shown in Fig.~\ref{fig:ds_sync} (a)), incurring considerable idle waiting as a result of the global barrier.

Thus, in the production network environment, taking BSP as the benchmark, an ideal synchronization scheme for distributed DNN training should achieve the following goals:
\begin{itemize}
\item \textbf{Communication efficiency:} it should be topology-aware and fully utilize network bandwidth while avoiding bottlenecks to reduce idle waiting of BSP in each iteration.
\item \textbf{Convergence accuracy:} it should maintain the same convergence accuracy as BSP in similar iterations, which is widely considered as the best in this aspect~\cite{DBLP:conf/ppopp/AwanHHP17,DBLP:conf/usenix/ZhangZXDHLHWXX17,DBLP:conf/infocom/BaoPCW20,DBLP:conf/sosp/PengZCBYLWG19}. 
\end{itemize}

Existing solutions, no matter whether system-level optimizations strengthening BSP (e.g., Ring~\cite{DBLP:conf/cluster/MamidalaLP04} or Hierarchical Allreduce~\cite{DBLP:conf/mlsys/ChoF0H19}) or algorithmic optimizations replacing BSP (e.g., asynchronous parallelism (ASP)~\cite{asp} or stale synchronous parallelism (SSP)~\cite{nips:ssp}), do not achieve the above two goals simultaneously (details in $\S$\ref{sec:drawback}). For the former one, while solutions like Ring~\cite{DBLP:conf/cluster/MamidalaLP04}, Tree~\cite{DBLP:conf/ipps/PatarasukY07}, or Hierarchical Allreduces~\cite{DBLP:conf/mlsys/ChoF0H19} explore decentralized collective methods to improve bandwidth utilization, they still have a long dependency chain that may block downstream communications if network bottlenecks exist. Furthermore, due to their global synchronization nature, these solutions can hardly avoid idle waiting. For the latter one, by relaxing the global synchronization barrier with ASP~\cite{asp} or SSP~\cite{nips:ssp}, they avoid or defer idle waiting of BSP in each iteration. However, these methods often incur a slower convergence speed, i.e., requiring a larger number of training iterations. Particularly, as network bottlenecks accumulate the staleness (iteration gap between the fastest and the slowest worker) on the same worker, it can bring in outdated and noisy gradients, leading to convergence inaccuracy.

To this end, we propose DS-Sync, a new divide-and-shuffle synchronization to achieve both communication efficiency and convergence accuracy simultaneously ($\S$\ref{sec:design}). The key idea of DS-sync is to divide workers into non-overlap groups of different sizes to synchronize independently and periodically shuffle workers among groups to reach global consensuses. According to the network topology and situations, DS-Sync generates a periodical pattern to divide and shuffle all workers. In every iteration, any worker optimizes its model with local gradients and then synchronizes and averages the model parameters with workers in the same group. In the next iteration,  workers are shuffled among groups in the generated pattern so that any worker synchronizes parameters with some unseen workers from different groups in the previous iteration. Fig.~\ref{fig:ds_sync} compares our DS-Sync with PS (widely used in BSP, ASP, and SSP) and the topology-aware Hierarchical Allreduce to illustrate the advantages of DS-Sync. 

DS-Sync improves communication efficiency with {\em dividing} and ensures convergence accuracy with {\em shuffling}. By taking network bottlenecks into account, it divides workers into different groups to minimize the maximum communication cost of all groups to reduce idle waiting. The insight is that the more workers to synchronize, the more communication traffic and the more latency ($\S$\ref{sec:sync_scale}). DS-Sync always keeps bottlenecked workers or links in the smaller group to speed up, while other regular ones form larger groups. Therefore, bottlenecked workers can catch up with others to reduce idle waiting and improve communication efficiency. DS-Sync maintains convergence via iterative shuffling, which is formally proven in $\S$\ref{sec:convergence}. DS-Sync carefully designs the shuffling pattern to ensure that any worker can directly or indirectly exchange information with all the groups during the shuffling period. Intuitively, local updates of all groups in any iteration are iteratively merged in the upcoming iterations. As a result, all workers can reach global consensus iteratively instead of global synchronization in every iteration. Furthermore, DS-Sync extends the stochastic weight average (SWA) to ensemble the diverse local models for better generalization performance.

We have implemented DS-Sync in the PyTorch framework and conducted comprehensive testbed experiments ($\S$\ref{sec:eval}). Our results show that DS-Sync can improve communication efficiency without any loss in convergence speed or accuracy. For example, compared to the Allreduce BSP, DS-Sync can achieve up to 94\% improvement in terms of end-to-end training time to reach the target accuracy. In the scenario of bandwidth contention, DS-Sync improves communication efficiency by up to 2X while maintaining the same accuracy as BSP with a similar number of iterations.

Overall, this paper makes three key contributions:
\begin{enumerate}
	\item We propose DS-Sync, a new divide-and-shuffle synchronization scheme, to achieve both communication efficiency and convergence accuracy simultaneously.     
	\item We prove that DS-Sync converges to the same accuracy of BSP with a similar number of iterations under the nonconvex and smooth conditions of DNN.
    \item We implement DS-Sync in PyTorch and validate its effectiveness through extensive testbed experiments.
 \end{enumerate}

\section{Backgrounds and Motivations}
\label{sec:background}
In this section, we first overview the network bottlenecks in production clusters and then discuss the drawbacks of existing solutions in handling them. 
  
\subsection{Network Bottlenecks in Training Cluster}
\begin{figure}[t]
  	\centering
  	\includegraphics[width=0.85\linewidth]{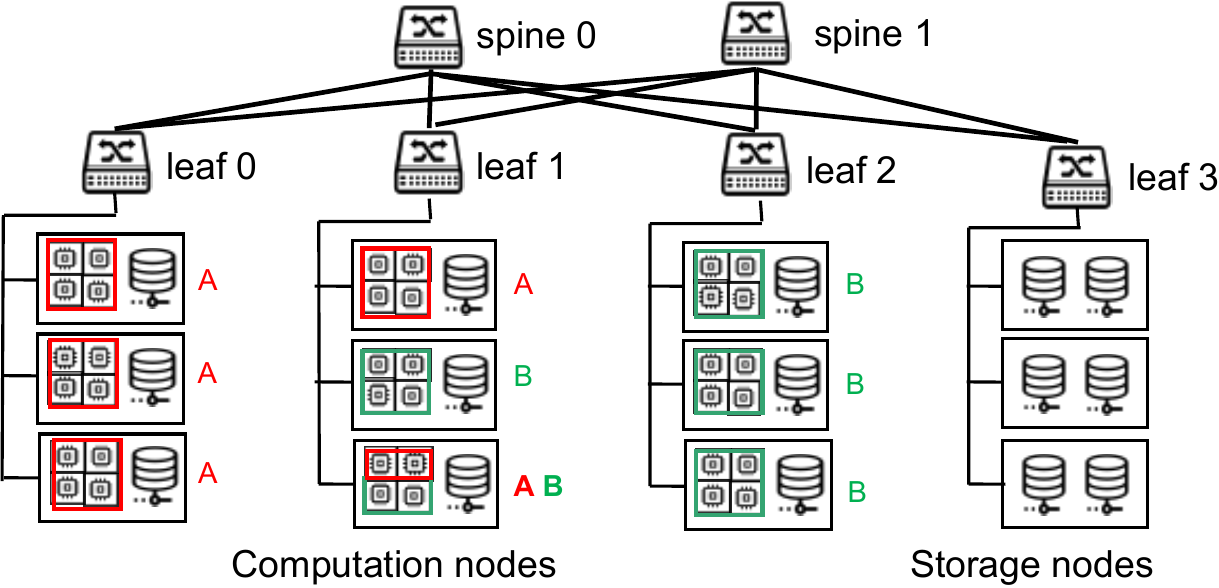}
  	\caption{An example of GPU cluster shared by tasks A and B. }
  \label{fig:gpu_cluster}
  \end{figure}

Fig.~\ref{fig:gpu_cluster} illustrates a typical architecture of a training cluster shared by different tasks, including computation nodes, storage nodes, and the network. 
The computation nodes have accelerators like GPUs or TPUs but with limited storage as the local cache.
The storage nodes with RAID work as the elastic and fast Network File System (NFS). 
The network usually adopts hierarchical physical topology like spine-leaf, which is scaled up easily by adding switches in each layer~\cite{DBLP:conf/ecms/BilalKKZHMMWC12}. 
All nodes are grouped into racks, each connected by a leaf switch. All leaf switches are connected to upper-level spine switches. 
In this setting, network bottlenecks may arise due to: 

\begin{enumerate}
	\item \textbf{Static topology heterogeneity:} Intra-rack communication is non-blocking, but inter-rack communication depends on the inter-rack link load and oversubscription ratio~\cite{DBLP:conf/osdi/ZhaoHYZYZYLWXW20}. 
	If a task has multiple inter-rack connections, it can suffer from the oversubscription problem.
	\item \textbf{Dynamic inter-rack bandwidth contention:} In production clusters, there are often background flows from other training tasks that compete for the inter-rack bandwidth~\cite{DBLP:conf/usenix/JeonVPQXY19,DBLP:conf/osdi/ZhaoHYZYZYLWXW20,DBLP:conf/mlsys/LuoWKCN20} (e.g. the leaf 1 in  Fig.~\ref{fig:gpu_cluster}). 
	Regular pairs of workers can deliver over 2x the throughput of those delayed by the inter-rack bandwidth contention~\cite{DBLP:conf/mlsys/LuoWKCN20}.
	\item \textbf{Dynamic end-host bandwidth contention:} Different distributed training tasks can co-locate in the same physical node to occupy different GPUs or TPUs but share the same NIC~\cite{DBLP:conf/usenix/JeonVPQXY19, DBLP:conf/osdi/ZhaoHYZYZYLWXW20} (e.g. the 3rd node of leaf 1 in Fig.~\ref{fig:gpu_cluster}). 
	This causes end-host bandwidth contention that potentially slows down the communication of corresponding workers.
\end{enumerate}

\subsection{Existing Solutions and Their Drawbacks}\label{sec:drawback}
BSP can be significantly influenced by the above network bottlenecks. In general, there are two categories of existing works to improve it. One is the system-level optimizations strengthening BSP~\cite{osdi:parameterserver, DBLP:conf/osdi/JiangZLYCG20,DBLP:conf/ppopp/AwanHHP17,DBLP:conf/cluster/MamidalaLP04, DBLP:conf/ipps/PatarasukY07,DBLP:conf/usenix/ZhangZXDHLHWXX17,DBLP:conf/infocom/BaoPCW20,DBLP:conf/sosp/PengZCBYLWG19,DBLP:conf/mlsys/ChoF0H19, DBLP:conf/mlsys/LuoWKCN20,DBLP:conf/apnet/XiaZZW0J019,DBLP:journals/corr/abs-2008-08445}, the other is algorithmic optimizations replacing BSP~\cite{asp,nips:ssp,DBLP:conf/ppopp/LiBGAH20,DBLP:conf/nips/LianZZHZL17, DBLP:conf/icml/AssranLBR19}. However, neither of them can achieve the aforementioned two goals simultaneously in production clusters.
\subsubsection{System-level Optimization} 
Due to its global synchronization nature, BSP can hardly eliminate the idle waiting of regular workers and links. 
Some system-level works employ various topologies like PS~\cite{osdi:parameterserver, DBLP:conf/osdi/JiangZLYCG20}, Ring~\cite{DBLP:conf/cluster/MamidalaLP04}, and double tree~\cite{DBLP:conf/ipps/PatarasukY07} to fully use network bandwidth.
Some others explore different underlying network optimizations, including overlapping communication and computation~\cite{DBLP:conf/ppopp/AwanHHP17,DBLP:conf/usenix/ZhangZXDHLHWXX17,DBLP:conf/infocom/BaoPCW20,DBLP:conf/sosp/PengZCBYLWG19,DBLP:journals/corr/abs-2008-08445,DBLP:conf/apnet/XiaZZW0J019}, 
RDMA~\cite{DBLP:conf/sigcomm/GuoWDSYPL16,DBLP:conf/sigcomm/YiXC017},
in-network aggregation~\cite{DBLP:journals/corr/abs-1803-01491,DBLP:conf/nsdi/LaoLMCWAS21,DBLP:conf/nsdi/SapioC0NKKKMPR21}, 
congestion control\cite{DBLP:conf/sigcomm/AlizadehGMPPPSS10,DBLP:conf/hotnets/ChenH0WT13}, 
flow scheduling\cite{DBLP:conf/nsdi/Bai0WCHT15,DBLP:conf/infocom/LiBCHZLY17,DBLP:conf/sigcomm/ChenL0L18}, 
and coflow scheduling\cite{DBLP:conf/sigcomm/ChowdhuryZS14,DBLP:conf/sigcomm/ZhangCY0CG16,DBLP:conf/infocom/Zhao00YTGZLW15}. 
However, all these system-level works are either topology-agnostic and/or contention-vulnerable.

Both heterogeneous topologies and dynamic contentions lead to network bottlenecks. 
Unaware of the network topology\footnote{Current Allreduce implementations like NCCL and MPI are unaware of physical network topology. NCCL only detects different physical link types within the node such as NVlink, PCI-E, and network.},
these works cannot be guaranteed to align the logical topology with the physical one. 
Multiple connections may cross and compete for the inter-rack links, resulting in the network bottleneck of oversubscription.
Additionally, Ring Allreduce and Tree Allreduce introduce long dependency in the logical topology and pipelines. 
They are vulnerable to bandwidth contentions on inter-rack links or end-host NIC since they can easily block downstream communications. 
Hence, the network bottleneck brings in significantly idle waiting time for BSP. 

Topology-aware Hierarchical Allreduces are still sub-optimal since the network bottleneck essentially stalls regular workers and links.
Hierarchical Allredcue~\cite{DBLP:conf/mlsys/ChoF0H19, DBLP:conf/mlsys/LuoWKCN20, DBLP:conf/apnet/WanZWHZ020} decomposes the global synchronization into sequential communications to localize the network bottleneck. 
Specifically, Blueconnect~\cite{DBLP:conf/mlsys/ChoF0H19} makes three serial communications steps, including intra-rack reduce, inter-rack allreduce, and intra-rack broadcast. 
The serial communications make all workers related with intra-rack communication wait for the inter-rack communication and vice versa. 
Plink~\cite{DBLP:conf/mlsys/LuoWKCN20} takes a further step to slice data into chunks and uses a pipeline to overlap inter-rack and intra-rack communications. 
However, some workers simultaneously participate in inter-rack and intra-rack communications, which is the bottleneck slowing down communications in other pipeline stages. 

\begin{figure}[b]
  	\centering
  	\includegraphics[width=0.85\linewidth]{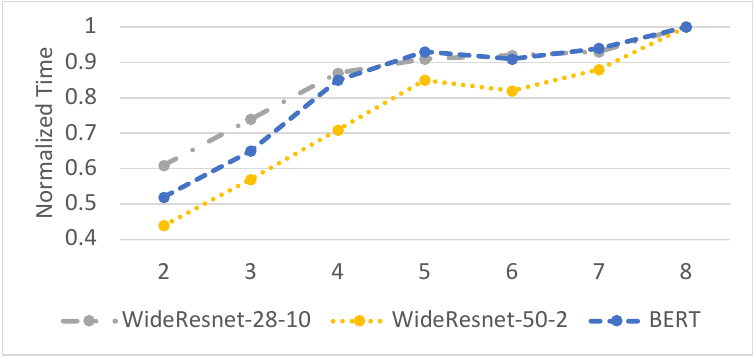}
  	\caption{ Normalized communication time vs. worker number (normalized by the maximum time value)}
  	\label{fig:sync_scale}
  \end{figure}

\label{sec:sync_scale}
Even in an ideal and uniform network, the BSP communication time increases with the total number of workers. Specifically, the communication cost of point-to-point transferring a model of S bytes can be modeled as $L \alpha+S \beta$~\cite{DBLP:conf/imw/SarvothamRB01}, where $\alpha$ denotes the latency, $\beta$ is the transfer delay for one byte, and $L$ is the number of times to invoke model synchronization layer by layer~\cite{DBLP:conf/ppopp/AwanHHP17,DBLP:conf/usenix/ZhangZXDHLHWXX17,DBLP:conf/infocom/BaoPCW20,DBLP:conf/sosp/PengZCBYLWG19} to overlap computation and communication. The more workers $N$ to synchronize, the more point-to-point communications and total communication traffic, as summarized in Tab.~\ref{tab:sync_cost} for popular BSP methods. We also conduct experiments to verify it in different worker numbers and different models, and Fig.~\ref{fig:sync_scale} shows the trend that communication time increases with worker numbers, which motivates us to divide all workers and put the bottleneck in the small enough one to speed up.
\begin{table}[t]
\centering
\caption{Communication cost comparison }
\begin{tabular}{|c|c|c|}
\hline
               & Latency    & Transfer Delay \\ \hline
PS             & $2(N-1)L\alpha$           & $2(N-1)S\beta / P$                         \\ \hline
Ring AR        & $2(N-1)L\alpha $ & $2(N-1)S\beta/N$                         \\ \hline
Double Tree AR & $2(\log N + k) L\alpha$    & $2(\log N+k) S\beta /k $                         \\ \hline
\end{tabular}
\label{tab:sync_cost}

$P$ stands for the PS number, and $k$ is the data chunk number.
\end{table}
  
\subsubsection{Algorithmic Optimization} 
Previous algorithmic works only focus on occasional stragglers due to stochastic events like system interruption. 
This approach offers more flexibility by relaxing the synchronization conditions, expecting the straggler to happen on another worker later.
However, workers delayed by the network bottleneck accumulate the staleness to a high level, which is beyond their expectations and results in convergence and throughput problems.
 
ASP~\cite{asp} does not have the guarantee to converge at all. It directly gets rid of all synchronizations and hopes that all workers have similar paces. However, the accumulated high-level staleness due to network bottlenecks is against it. Although ASP has no waiting time, the consistent outdated gradients due to network bottlenecks on the same worker usually result in lower convergence accuracy.

Due to accumulated staleness, SSP~\cite{nips:ssp,DBLP:conf/ppopp/LiBGAH20} suffers from convergence inaccuracy and communication inefficiency. 
It defers global synchronization until exceeding staleness bound. 
If the bound is high, the consistent staleness leads to a slower convergence rate and even failures sometimes~\cite{DBLP:conf/iclr/DaiZDZX19}. 
Otherwise, the staleness also frequently invokes global synchronization with low bound, resulting in idle waiting, just like BSP.

Originated from IoT network, Gossip~\cite{DBLP:conf/nips/LianZZHZL17, DBLP:conf/icml/AssranLBR19,wang2019matcha} suffers from inefficient point-to-point communications and unawareness of network bottlenecks. 
It breaks the global barrier into chained local barriers. 
Every worker only sends and receives messages to the neighbors in the sparsely connected graph to relay information iteratively. 
Gossip has the congestion problem in the end host if there are over two peers in a worker's neighborhood\cite{wang2019matcha}. 
The connected graph of Gossip can be time-variant, such as the dynamic exponential graph\cite{DBLP:conf/icml/AssranLBR19} and random matching\cite{wang2019matcha}. 
Hence, there can be many connections to worsen the inter-rack contention in some iterations. 
Furthermore, the longer chain of gossip propagation, the slower it is to reach consensus~\cite{DBLP:conf/icml/AssranLBR19}.
\section{Design}
\label{sec:design}

This section presents the detailed design of DS-Sync. First, we introduce the overall workflow of DS-Sync. Then, we discuss how workers are divided and shuffled to handle different network bottlenecks.

\subsection{Overall Workflow}

\begin{lstlisting}[language=Python, name=Code ,caption=The workflow of DS-Sync, float=t, label={list:workflow}]
import DS_Sync
def train(dataloader, i=rank):
	#detect the physical topology and initialize the group list for DS-Sync accordingly
	topology = DS_Sync.topologyDetect() 
	groupList = DS_Sync.initGroups(topology)
	#initialize all wokrers' models in the same way
	model, lossCriteron, optimizer, SWA_model = DS_Sync.model_generator()
	for t in range(Total_iteration):
		#adjust the grouping according to dynamic bandwidth variations in every 100 iterations
		if t%period==0 And bandwidth changes:
			groupList = DS_Sync.adjustGroups(groupList)
		#FP and BP computation on the sampled data batch
		x, y = DataLoader()
		pred = model(x)
		loss = LossCriteron(pred, y)
		gradient = loss.backward()
		#locally update model parameters 
		optimizer.step()
		optimizer.zero_grad()
		#pick the proper group to synchronize for worker i in the iteration t
		group = groupList[t%len(groupList)][i]
		#synchronize and average the parameters within the same group and record bandwidth
		model = DS_Sync.Average(model, group)
		#track and record communication cost for the group adjustment
		DS-Sync.commRecord()
		#local stochastic weight average ensemble
		SWA_model = factor*SWA_model+ (1-factor)*model
	#weight average all local SWA_models to ensemble all trajectories as the final model to test
	Final_model = AverageAllreduce(model, world)
\end{lstlisting}
At its heart, by taking network bottlenecks into account, DS-Sync periodically divides and shuffles workers to form multiple non-overlapping groups of different sizes to synchronize independently. DS-Sync first senses the physical network topology and bandwidth contentions, and then it generates the periodical divide-and-shuffle pattern accordingly (details in $\S$\ref{sec:ds_algo}).  DS-Sync synchronizes and averages workers' model parameters in the same group in every iteration according to the divide-and-shuffle pattern. DS-Sync also extends SWA to ensemble diverse trajectories of different workers. DS-Sync can be easily integrated with the DL framework like PyTorch as stated in Pseudo Code~\ref{list:workflow}. Specifically, DS-Sync has the following key steps:
\begin{enumerate}
	\item \textbf{Topology Detection} \texttt{topologyDetect()}: While the network topology can be known to the operators, it is often unknown to the general users. Fortunately, network profiling such as DPDK and iPerf can be employed to measure the round-trip latency and bandwidth between worker pairs. The topology information can thus be derived from the measurements~\cite{DBLP:conf/mlsys/LuoWKCN20}.
	\item \textbf{DS-Sync Group Initialization} \texttt{initGroups()}:   According to the static topology heterogeneity, DS-Sync initializes the periodical divide-and-shuffle pattern that divides workers into the inter-rack group and intra-rack groups as described in $\S$\ref{sec:static}. In this way, DS-Sync can minimize its own connections and communication traffic crossing the inter-rack links to avoid oversubscriptions. 
	\item \textbf{DS-Sync Group Adjustment} \texttt{adjustGroups()}: Once sensing bandwidth contention due to background flow changes in a period, DS-Sync further adjust the periodical divide-and-shuffle pattern. 
	Since the bottlenecked group is caused by bandwidth contention on either inter-rack link or end-host NIC, the related inter-rack or intra-rack group can be further divided into smaller ones. 
	Therefore, DS-Sync decreases the max group communication cost and keeps all groups in similar paces to proceed.(details in $\S$\ref{sec:dynamic_link}~\ref{sec:dynamic_nic} ). 
	\item \textbf{Parameter Synchronization} \texttt{code line 12-24}: 
	DS-Sync synchronizes all groups in parallel. 
	Within the group, it uses Allreduce to average the parameters of workers in the group after they locally optimize their model parameters.
	The model parameter contains all information about the past parameter updating steps. 
	In the next iteration, DS-Sync shuffles workers among groups so that different groups can exchange the past updating information with each other. 
	Specifically, the model parameter of worker i in iteration T is $w^{(i)}_T=w_0+\sum_{t=0}^T \prod_{k=t}^T W_k e_i \bigtriangleup_t$, 
	where $\bigtriangleup_t$ is local update vector of all workers in the iteration t, $W_k$ is the group average in the iteration K, $e_i$ is a unit vector that only the i-th element is 1, and $\prod_{k=t}^T W_k e_i \bigtriangleup_t$ is the approximated distributed average. 
	DS-Sync also tracks every workers' communication cost to adjust group patterns later. 
	\item \textbf{Extended SWA Ensemble} \texttt{code line 25-28}: DS-Sync can further exploit the diverse parameter trajectories of different workers to improve test accuracy by ensembling. Every worker applies exponential moving average on local parameters in every iteration to get a local SWA model~\cite{DBLP:conf/uai/IzmailovPGVW18}. At the end of the training, all workers average their SWA models to ensemble all trajectories to reach a more flatten optimal solution. The flatten optimal solution is more robust to small data perturbation than a sharp one, which generalizes better on unseen test data~\cite{DBLP:conf/iclr/KeskarMNST17}.
\end{enumerate}
	
	DS-Sync improves communication efficiency by dividing workers into groups of different sizes and guarantees convergence with shuffling workers among groups to reach consensus iteratively. 
	According to the network situation, DS-Sync always keeps bottlenecked workers or links in smaller groups, while other regular ones form larger groups. 
	Then delayed workers or links have less latency and communication traffic than others. In this way, bottlenecked workers in small groups can catch up and reduce the idle waiting of others. 
	Additionally, DS-Sync shuffles workers among groups in every iteration. 
	It follows the periodical divide-and-shuffle pattern guaranteeing that the past local updates of different workers can be merged by iterative propagation. 
	Therefore, any worker can directly or indirectly get the past global updating information iteratively instead of immediately like BSP. 
	Finally, all workers can reach global consensus iteratively and maintain the convergence on training data (formal proof in $\S$~\ref{sec:convergence}). 
	DS-Sync also extends SWA to ensemble all diverse local models for better generalization on test data.

\subsection{Handling Different Network Bottlenecks}
\label{sec:ds_algo}
Now, we introduce how DS-Sync periodically divides and shuffle workers into groups according to different network bottlenecks. To achieve both goals, there are two principles for generating the divide-and-shuffle group pattern:
\begin{itemize}
 \item It should decouple network bottlenecks from others and keep them in a smaller group to alleviate bottlenecks.
 \item It should guarantee that all workers can directly or indirectly exchanges information by shuffling. 
 \end{itemize}
DS-Sync generates divide-and-shuffle group patterns in the group initialization and adjustment. In the initialization, DS-Sync forms inter-rack and intra-rack groups according to the static network topology. Then during the training, it reacts to the dynamic bandwidth contention by further dividing the bottlenecked group into multiple smaller groups. Specifically, DS-Sync handles three kinds of network bottleneck as follows:

\subsubsection{Static Topology Heterogeneity}
\label{sec:static}
\begin{figure}[t]
  	\centering
  	\includegraphics[width=0.9\linewidth]{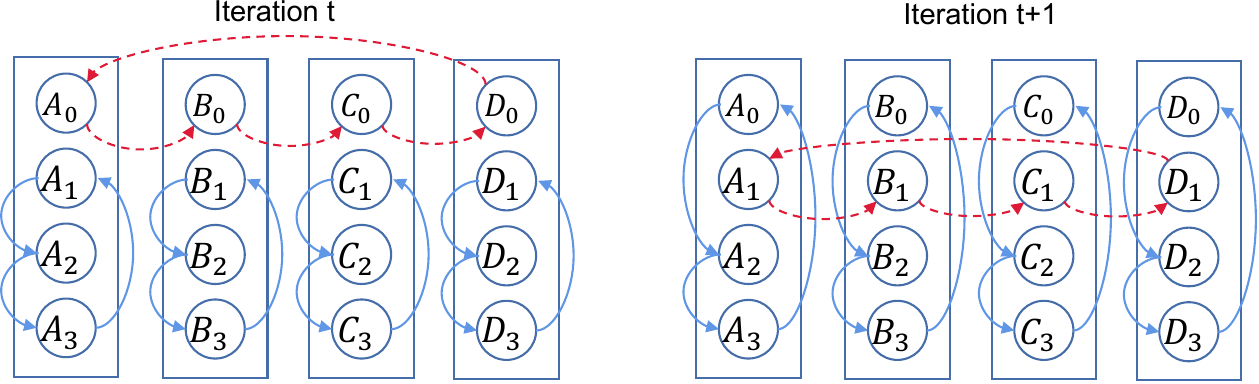}
  	\caption{Group initialization divides all workers into inter-rack and intra-rack groups and shuffles workers to be the representative in the inter-rack group by turns.}
  \label{fig:static}
 \end{figure}
 After detecting the physical network topology in \texttt{topologyDetect()}, DS-Sync initializes the group pattern in \texttt{initGroups()} by dividing workers into the inter-rack groups and intra-rack groups. Firstly, all workers in the same rack form intra-rack groups. DS-Sync further separates one worker from each intra-rack group to be the representative forming an inter-rack group. In the shuffling, all workers in the same rack take turns to be the representative.
Since Allreduce is used for synchronization within any group, DS-Sync can guarantee itself has only one connection crossing one inter-rack link. Therefore, it is free of the oversubscription problem caused by itself. Every intra-rack group exchanges members with the inter-rack group to reach the global consensuses iteratively. Fig.~\ref{fig:static} illustrates an example of four racks how workers are divided and shuffled in inter-rack and intra-rack groups.

\subsubsection{Inter-rack Link Bandwidth Contention}
\begin{figure}[b]
  	\centering
  	\includegraphics[width=0.9\linewidth]{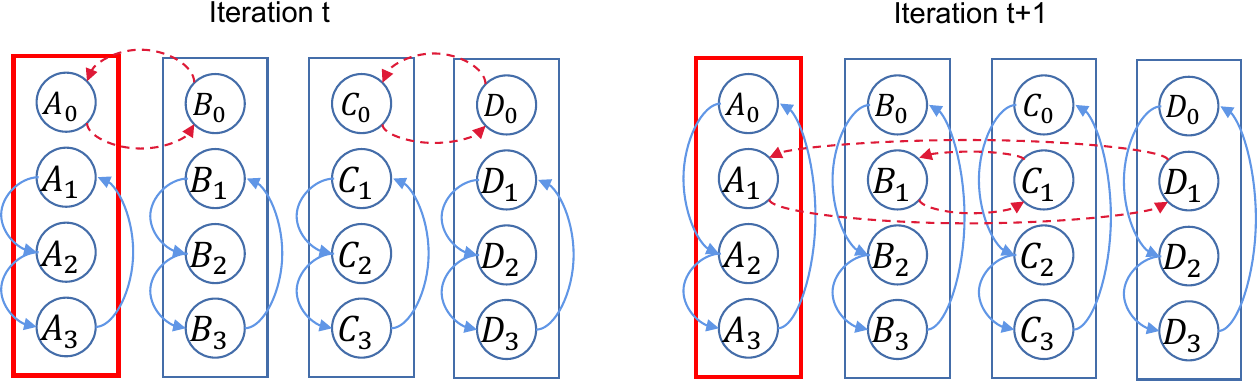}
  	\caption{Rack A (in the bold and red rectangles) has some background flows from other tasks to share inter-rack bandwidth. DS-Sync further divides inter-rack groups to reduce the group size in group adjustment during training.}
  \label{fig:dynamic}
 \end{figure}
\label{sec:dynamic_link}
 If any rack is sensed to have slow inter-rack communication due to background flows from other tasks in the same rack, DS-Sync adjusts the inter-rack group in \texttt{adjustGroups()}. It further divides the inter-rack group and keeps the influenced representative in the smallest groups. Specifically, DS-Sync put each delayed representative with another regular representative to form an inter-rack group of two workers, while other regular representatives form a large inter-rack group. In every iteration, the regular representative in the small inter-rack group is exchanged with another representative in the large inter-rack group. Workers in the same rack also take turns to be the representative to get global information exposed. Fig.~\ref{fig:dynamic} shows an example that the inter-rack group is further divided into two inter-rack groups.

\subsubsection{End-host NIC Bandwidth Contention}
\label{sec:dynamic_nic}
\begin{figure}[t]
  	\centering
  	\includegraphics[width=0.95\linewidth]{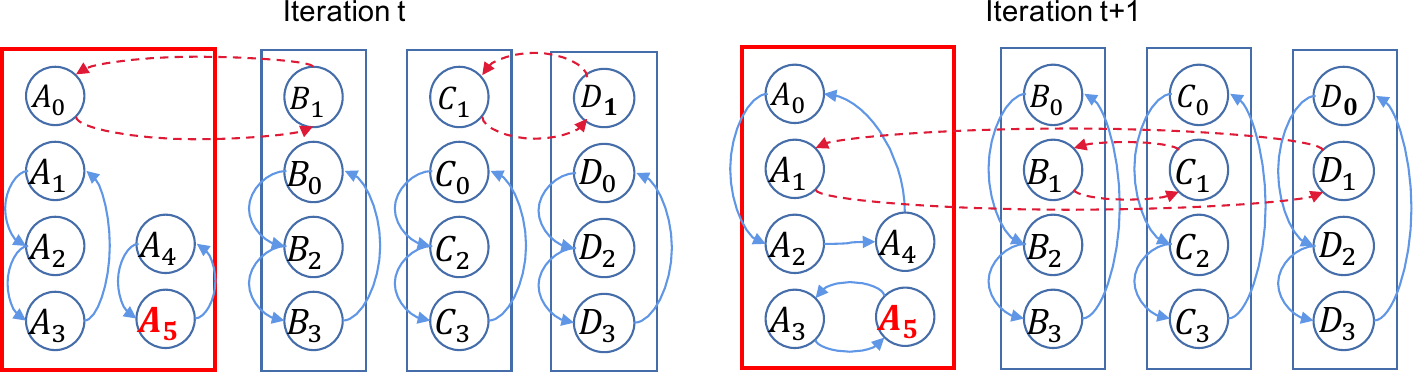}
  	\caption{In rack A, worker $A_5$ (in the bold and red characters) has another task sharing the end-host NIC. The upper-level link of rack A also has smaller bandwidth due to background flows.  Besides the inter-rack group, DS-Sync further divides the intra-rack group to alleviate the bottleneck on worker $A_5$. }
  \label{fig:larger}
 \end{figure}
If DS-Sync finds a worker has smaller bandwidth due to sharing the end-host NIC with another task in the same node, \texttt{adjustGroups()} also adjust the group pattern accordingly. DS-Sync always keeps the delayed worker in the intra-rack group. If there are enough workers in the rack hosting the delayed worker, DS-Sync further divides the intra-rack group into sub ones. Similar to $\S$\ref{sec:dynamic_link}, DS-Sync pair the delayed worker with another regular worker to form a small intra-rack group, and the rest workers form a large intra-rack group. The large intra-rack group exchanges workers with the small group as well as the inter-rack group. Therefore, the large group works as an information hub of the rack so that all subgroups can exchange information. 
\section{Analysis}
\begin{table}[b]
\centering
\caption{Communication Cost Summary}
\label{tab:comm_summarization}
\begin{tabular}{|c|c|c|}
\hline
Methods       & Latency                  & Transfer Delay                                       \\ \hline
PS & $2(N-1)L\alpha$              & $\frac{2[(C-1)GS\beta_{2}+(G-1)S\beta_{1}]}{P}$ \\ \hline
Ring          & $2(N-1)L\alpha$              & $2(N-1)S\beta_{2}/N$                            \\ \hline
Double Tree   & $2(logN+k)L\alpha$           & $2(LogN+k)S\beta_{2}/k$                         \\ \hline
Hierarchical  & $2(G+C-2)L\alpha$     & $\frac{2(G-1)S\beta_{1}}{G}+\frac{2(C-1)S\beta_{2}}{C}$      \\ \hline
Gossip        & $2L\alpha$                   & $\geq 4S\beta_{2}$                                   \\ \hline
DS-Sync       & $2L\alpha$ or $2L(G-1)\alpha$ & $2S\beta_{2}$ or $\frac{2(G-2)S\beta_{1}}{G-1}$        \\ \hline
\end{tabular}

DS-Sync keeps the bottlenecked links and related workers in the smallest group of two workers, and $G-1$ is the size of the intra-rack group. 
\end{table}

In this section, we start with the quantitative communication analysis on DS-Sync and other methods in the scenario of the inter-rack bottleneck. 
Then, we mathematically prove that DS-Sync has the same convergence accuracy and rate as BSP in the nonconvex and smooth conditions of DNN.

\subsection{Communication Time Analysis}
For simplicity, we analyze the communication time in a simplified and flattened hierarchical network. 
A total number of N workers are evenly allocated into C racks, and each rack has G nodes. 
We assume that the same background flows of other tasks interfere with the available inter-rack bandwidth. 
$\alpha$ is the latency for one hop, $\beta_1$ denotes intra-rack transfer delay for one byte, and $\beta_2$  strands for inter-rack transfer delay for one byte. 
We also assume that all methods perfectly align their logical topologies with the physical network topology to minimize their inter-rack connections and communication traffic. 
The DNN model is synchronized for $L$ times to overlap communication and computation. 
	
Tab~\ref{tab:comm_summarization} summarizes the communication cost consisting of latency and transfer delay for various previous methods and our DS-Sync. 
PS is widely used in BSP, ASP, and SSP, but it suffers from congestion on the inter-rack link because of high crossing rack traffic. 
Due to the long dependency in logical topologies and pipelines, Ring Allreduce and Tree Allreduce are vulnerable to the contention of background flows. 
If an inter-rack link is bottlenecked, it slows down communication in the other stages. 
Hierarchical Allreduce has multiple serial communication steps so that the communication cost is the sum of all steps. 
If it further slices data into chunks to transfer in the pipeline, the workers in both inter-rack and intra-rack levels slow down the whole pipeline due to simultaneously sending and receiving messages in different stages. 
In Gossip algorithms, each rack has two or even more inter-rack connections in some iterations for the time-varied graph. 
Any worker also suffers from the in-cast problem if it has multiple neighbors. 
Due to the dependency of chained local barriers, bottlenecked workers finally stall others.

In contrast, our DS-Sync divides all workers into multiple non-overlap groups to synchronize in parallel. 
According to the network topology and situations, DS-Sync always assigns bottlenecked workers and links in smaller groups to catch up. 
The total communication time of DS-Sync is determined by the max one of all inter-rack and intra-rack groups.

\subsection{Convergence Analysis}
\label{sec:convergence}
We mathematically prove that DS-Sync converges properly in nonconvex and smooth conditions of DNN. 
Firstly, we state some common assumptions of nonconvex and smooth DNN. 
Then we prove how DS-Sync can achieve global consensuses on model parameters iteratively. 
Finally, we show that DS-Sync can converge properly in nonconvex and smooth conditions.

\subsubsection{Assumptions}
Following the previous convergence theory for SGD in nonconvex and smooth conditions~\cite{DBLP:journals/siamrev/BottouCN18}, 
we make some common assumptions in the optimization community for DNN as follows:
\begin{assumption}
\label{smooth}
\textbf{Lipschitzian smooth:} Any local function of worker $i$ $F_i(\cdot)$ is with L-Lipschitzian gradients.
\[
\left\|\bigtriangledown F_i(x;\xi)-\bigtriangledown F_i(y;\xi)\right\| \leq L\left\|x-y\right\|  
\]
\end{assumption}
\begin{assumption}
\label{bound}
\textbf{Bounded variance:} Assume the variance of stochastic gradient $\mathbb{E}_{i \sim \mathcal U([n])}E_{\xi\sim \mathcal D_i}\left\|\bigtriangledown F_i(x;\xi)-\bigtriangledown f(x)\right\|^2$ is bounded for any parameters $x$ with worker $i$ uniformly sampled from $\{1, \cdots, n\}$ and data batch $\xi$ from the distribution $\mathcal D_i$. This implies there exist constants $\sigma$ such that
\[
\mathbb{E}_{i \sim \mathcal U([n])} \\ E_{\xi\sim \mathcal D_i}\left\|\bigtriangledown F_i(x;\xi)-\bigtriangledown f(x)\right\|^2 \leq \sigma^2
\]
\end{assumption}

\subsubsection{Consensus Proof}
Before formally analyzing the convergence, we first prove that DS-Sync can reach the global consensus for the distributed average problem iteratively. 
In distributed average problem, each node $i$ starts with a number $x_{0}^{(i)}$, 
and $X_k$ stands for the vector $[x_{k}^{(0)},x_{k}^{(1)},...,x_{k}^{(N)}]^T$ in the iteration k, 
and the $W_k$ is a N-by-N transition matrix in $X_{k+1}=W_kX_k$, which indicates how workers exchange parameters in the iteration k. 
The goal of distributed average is to approximate the average $\bar{x}=\frac{1}{N}\sum_i x_{0}^{(i)}$ in every worker by K iterations as $X_K = W_{K-1}W_{K-2}\cdots W_{0}X_0$.

In DS-Sync, any $W_k$ represents how workers are divided into multiple non-overlap groups to be synchronized and averaged in the iteration k. 
The entry $w_{i,j}$ is either 1/group size or $0$, which means worker i and j are in the same group or not. 
The $W_k$ is the symmetric doubly stochastic matrix, in which every entry is not negative, and the sum of any row or column is one. 
Unlike the transition matrix of the Gossip representing a sparsely connected graph, the $W_k$ of DS-Sync stands for multiple fully connected sub-graphs covering all nodes without overlapping nodes and edges, which decouples network bottlenecks from others. 
Furthermore, $W_k$ follows the periodical divide-and-shuffle pattern such that $W_k = W_{k+B}$ and $\prod_{i=0}^{B}W_i$ indicates a connected graph of all nodes. 
It can be easily calculated that the second largest eigenvalue $\rho := \lambda_2(\prod^{(l+1)B-1}_{k=lB}W_k) < 1$.  
According to previous works in distributed average~\cite{DBLP:journals/pieee/NedicOR18},  the worst-case rate of convergence can be related to the second-largest as stated in Lemma~\ref{lemma_1}.

\begin{lemma}
\label{lemma_1}
We can bound the average error in worker i for DS-Sync as
\[
	\left\|\frac{\mathbf{1}_N}{N}X_0-\prod_{k=0}^{kB} W_{k}e_iX_0\right\|^2 \leq \rho^k\left\|X_0\right\|^2 \quad \ \forall i \text{ and } k\geq B
\]
\end{lemma}
\subsubsection{Convergence Proof}
The formal proof of DS-Sync convergence is based on the iteratively reached consensus on historical updates and the similarity of recent updates from smooth assumptions of DNN.
The model parameter of the worker can be rewritten as  $w^{(i)}_T=w_0+\sum_{t=0}^T \prod_{k=t}^T W_k e_i \bigtriangleup_t$, 
in which $\bigtriangleup_t$ is local update vector of all workers in the iteration t and $\prod_{k=t}^T W_k e_i \bigtriangleup_t$ is its approximated average. 
According to Lemma~\ref{lemma_1}, the approximated average in worker i for the historical update $\bar{\bigtriangleup}_t$ is bounded as $\left\|\bar{\bigtriangleup}_t-\prod_{k=t}^{T} W_{k}e_i\bigtriangleup_t \right\|^2 \leq \rho^{(T-t)/B}\left\|\bigtriangleup_t \right\|^2 $. 
DS-Sync can achieve better global consensus on earlier updates during the training. 
Although the recent update information may not be approximated very well, the smooth conditions of DNN bound the variances of gradients from different workers. 
Intuitively, different workers have slight differences in their local model parameters since most early past updating information has been averaged properly. 
According to the assumption of Lipschitzian smooth and bounded variance, the similar model parameters of different workers should have similar expectations of stochastic gradients on randomly sampled data batches. 
Therefore, we can formally prove the convergence of DS-Sync in the following theorems.
\begin{theorem}
\label{therom_1}
(Convergence of DS-Sync). Based on Assumption~\ref{smooth} and~\ref{bound} and Lemma~\ref{lemma_1}, we show that the gradients of all workers converge to be 0 with the same rate $O(1/\sqrt{K})$ as BSP~\cite{DBLP:journals/siamrev/BottouCN18}. In other words, DS-Sync can reach a minimal point in the nonconvex and smooth case like BSP. Specifically, if the total number of iterate K is sufficiently large, then it is bounded as follows: 
\begin{small}
\[
\frac{\sum_{k=0}^{K-1}\mathbb{E}\left\|\bigtriangledown f(\frac{X_k\mathbf{1}_N}{N}) \right\|^2} {K} 
\leq \frac{8(f(0)-f^*)L}{K}+\frac{(8f(0)-8f^*+4L)\sigma}{\sqrt{Kn}}
\]
\end{small}
\end{theorem}
\section{Evaluation}
\label{sec:eval}
In this section, we first introduce how DS-Sync and baselines are implemented as well as the experiment settings. Then, we discuss the experimental results of DS-Sync in various scenarios with network bottlenecks to seek answers to the following questions:

\begin{enumerate}
\item \textbf{Can DS-Sync achieve end-to-end performance advantages in various scenarios?} Our extensive evaluation verifies that DS-Sync has up to 94\% improvements over the baselines in terms of the end-to-end training time in different network bottleneck scenarios.
\item \textbf{Can DS-Sync alleviate different network bottlenecks to improve communication efficiency?} In the three network bottleneck scenarios, DS-Sync consistently achieves minimal communication time, which improves the efficiency by up to 2x over BSP baselines. 
\item \textbf{Can DS-Sync maintain a similar convergence rate and accuracy as BSP?} Different from ASP or SSP, DS-Sync can achieve the same accuracy as BSP with a similar number of iterations in the experiments. Especially for the smaller dataset, DS-Sync can have slightly higher accuracy due to the SWA ensemble. 
\end{enumerate}

\subsection{Implementation and Experiment Settings}
\label{sec:exp_setting}

\subsubsection{Implementation}
We implement DS-Sync and integrate it (as well as BSP baselines, including Allreduce BSP, Hierarchical Allreduce BSP, ASP (can only use PS~\cite{osdi:parameterserver}), and Gossip (dynamic exponential graph)) with PyTorch. We choose the NCCL library as the communication backend for all methods and leave itself to determine the proper Allreduce topologies (Tree by default) for BSP baselines and group synchronization of DS-Sync. Since ASP cannot use collective operators, we implement it with send and receive primitives. We register the backward hook and pre-forward hook for every layer to overlap communication and computation. In the backward hook, we conduct a local optimizer update and invoke the non-block synchronization layer by layer. In the pre-forward hook of the corresponding layer, we set the barrier waiting for the synchronization event to make sure updating is completed before forwarding computation. 

\subsubsection{Evaluation Metrics} We evaluate the following metrics to verify that DS-Sync can simultaneously achieve the two goals mentioned earlier: 
\begin{itemize}
\item \textbf{Time-to-Accuracy (TTA):} We use it to measure end-to-end training time. The target accuracy is set to the lowest one of BSP in five times running. We omit ASP results due to its failure to reach the target accuracy.
\item \textbf{Communication Time:} We measure the communication efficiency as the time from invoking the first non-block communication to all communications completed (For ASP, we only report the fastest worker). 
\item \textbf{Iteration Number:} We use the iteration number of reaching the target accuracy to measure the convergence rate. We omit it for ASP again for the same reason.
\item \textbf{Best Accuracy:} We record the best accuracy or F1 during the training to measure the convergence accuracy. Except for ImageNet, we run all experiments five times to get the mean and standard variation of best accuracy.
\end{itemize}

\subsubsection{Models and Training Settings}
We evaluate \textit{WideResnet-28-10}~\cite{DBLP:conf/bmvc/ZagoruykoK16} on  \textbf{CIFAR10/100}~\cite{cifardataset}, \textit{WideResnet-50-2}~\cite{DBLP:conf/bmvc/ZagoruykoK16} on \textbf{ImageNet}~\cite{DBLP:conf/cvpr/DengDSLL009}, and \textit{BERT}~\cite{DBLP:conf/naacl/DevlinCLT19} on \textbf{SQuADv1.1}~\cite{DBLP:conf/emnlp/RajpurkarZLL16}  in the fields of Computer Vision and Natural Language Processing. In principle, we follow the common practices~\cite{DBLP:conf/bmvc/ZagoruykoK16,DBLP:conf/naacl/DevlinCLT19} in the machine learning (ML) community to split data,  pre-process data, and set hyper-parameters. For DS-sync and all baselines, all hyper-parameters are the same as the original ML model papers except that we use the warm-up learning rate schedule for the large overall batch size. Due to the limited space, we do not list the detailed hyper-parameters here. 

\subsubsection{Testbed}
We use a private cluster of 10 GPU nodes as the experiment environment. Each physical node has NVIDIA V100 GPUs, Intel Xeon Silver 4114 CPU, 191 GB memory, and a NIC. All nodes are divided into two racks and connected in the spine-leaf network. The network has 20Gb Ethernet per link, and its oversubscription rate is 1. 

\subsubsection{Network Bottleneck Settings}
We manually align logical topologies of all baselines with the physical one in the best way to minimize inter-rack connections. We can run another background task training the BERT model distributedly on two workers of 2 racks to create background flows sharing the inter-rack link and end-host NIC. Specifically, we conduct experiments with the following different network bottlenecks:
\begin{itemize}

	\item \textbf{Inter-rack and end-host bandwidth contentions:} The background task has one worker in the same node but not the same GPU with our target task, which shares both the inter-rack and end-host bandwidth. Our task has eight workers in different nodes of two racks, including worker 0-4 in rack 0 and worker 5-7 in rack 1. One worker of the background task co-locates in the same physical node with worker 4, while the other worker is in rack 1.
	\item  \textbf{Inter-rack bandwidth contentions:} The background task has workers in the same rack but not the same node with our target task, which only shares the inter-rack bandwidth. Our task divides workers evenly into two racks: worker 0-3 in rack 0 and worker 4-7 in rack 1. The background task has two workers in the rest nodes of the two racks. 
	\item \textbf{Static topology heterogeneity:} There is no other background task causing bandwidth contention but only static topology heterogeneity.  Our task still divides workers evenly into two racks, including worker 0-3 in rack 0 and worker 4-7 in rack 1.
\end{itemize}

\subsection{Experiment Results}

 The experiment results verify that DS-Sync has the best communication efficiency without any loss in convergence rate and accuracy in various scenarios of different network bottlenecks.
 
\subsubsection{Inter-rack and End-host Bandwidth Contention}
\begin{figure}[t]
	\centering 
	\includegraphics[width=0.95\linewidth]{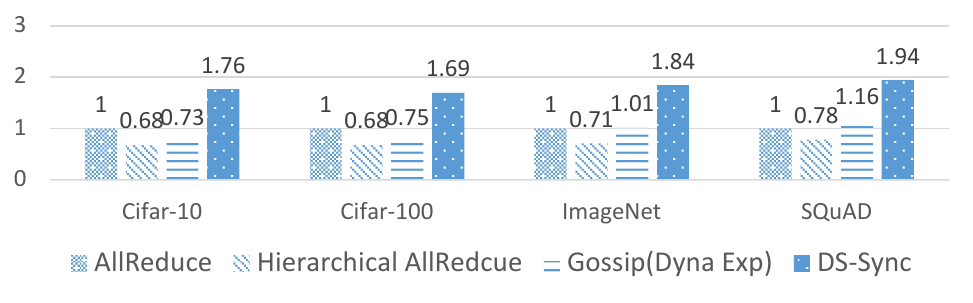}
	\caption{Speed up on TTA in inter-rack and end-host bandwidth contention}
	\label{fig:scenario_1}
\end{figure}
\begin{table}[t]
\caption{Summary on communication efficiency and convergence in inter-rack and end-host bandwidth contention}
\label{tab:scenario_1}
\begin{tabular}{|c|cccc|}
\hline
Dataset                   & Methods  & Comm. Time(ms)       & Iter. \# & Best Acc.        \\ \hline
\multirow{5}{*}{Cifar10}  & AR       & 540.1                & 6560     & 92.51$\pm$0.92\% \\
                          & Hier. AR & 790.2                & 6560     & 92.51$\pm$0.92\% \\
                          & ASP(PS)  & \textgreater{}567.3  & -        & 87.46$\pm$1.31\% \\
                          & Gossip   & 772.8                & 6360     & 92.47$\pm$1.17\% \\
                          & DS-Sync  & \textbf{338.1}       & 5640     & 92.62$\pm$0.56\% \\ \hline
\multirow{5}{*}{Cifar100} & AR       & 545.6                & 12740    & 76.79$\pm$0.28\% \\
                          & Hier. AR & 801.5                & 12740    & 76.79$\pm$0.28\% \\
                          & ASP(PS)  & \textgreater{}573.8  & -        & 66.23$\pm$2.03\% \\
                          & Gossip   & 778.3                & 12130    & 76.66$\pm$0.32\% \\
                          & DS-Sync  & \textbf{341.8}       & 11460    & 76.76$\pm$0.35\% \\ \hline
\multirow{5}{*}{ImageNet} & AR       & 825.7                & 341,664  & 76.27\%          \\
                          & Hier. AR & 1078.2               & 341,664  & 76.31\%          \\
                          & ASP(PS)  & \textgreater{}690.2  & -        & 72.72\%          \\
                          & Gossip   & 834.5                & 341,664  & 76.10\%          \\
                          & DS-Sync  & \textbf{412.5}       & 341,664  & 76.32\%          \\ \hline
\multirow{6}{*}{SQuAD}    & Methods  & Comm. Cost(ms)       & Iter. \# & Best F1          \\ \cline{2-5} 
                          & AR       & 3260.7               & 7,400    & 92.66$\pm$0.08\% \\
                          & Hier. AR & 4213.2               & 7,400    & 92.66$\pm$0.08\% \\
                          & ASP(PS)  & \textgreater{}2741.7 & -        & 89.85$\pm$1.21\% \\
                          & Gossip   & 2805.5               & 7,400    & 92.61$\pm$0.1\%  \\
                          & DS-Sync  & \textbf{1636.6}      & 7,400    & 92.64$\pm$0.05\% \\ \hline
\end{tabular}
\end{table}
 In this scenario,  DS-Sync first divides workers into inter-rack and intra-rack groups in the initialization. During the training, it further divides the intra-rack group in rack 1 into two smaller ones. DS-Sync keeps the influenced worker 4 in the group of 2 workers. By measuring the TTA, DS-Sync achieves the best end-to-end performance, as shown in Fig.~\ref{fig:scenario_1}.

Furthermore, we summarize all detailed metrics on communication efficiency and convergence for all methods in Tab.~\ref{tab:scenario_1}. Results show that DS-Sync gets the best communication efficiency by keeping inter-rack and end-host bottlenecks in the smallest independent groups. Meanwhile, it does not sacrifice the convergence rate and accuracy compared with BSP methods in all datasets. 

However, other baselines suffer from communication inefficiency or convergence inaccuracy. The smaller one of inter-rack and end-host bandwidth is the bottleneck for the whole Allreduce pipeline. Since Hierarchical Allreduce does not handle the contention in end-host NIC, its serial communications are even slower than Allreduce. The PS architecture of ASP and the dynamic exponential graph of Gossip can have multiple inter-rack connections, which worsens the inter-rack network bottlenecks. Furthermore, ASP fails to reach the target accuracy because accumulated high staleness brings in consistent outdated gradients harmful to the convergence.

\subsubsection{Inter-Rack Bandwidth Contention}
\begin{figure}[t]
	\centering 
	\includegraphics[width=0.95\linewidth]{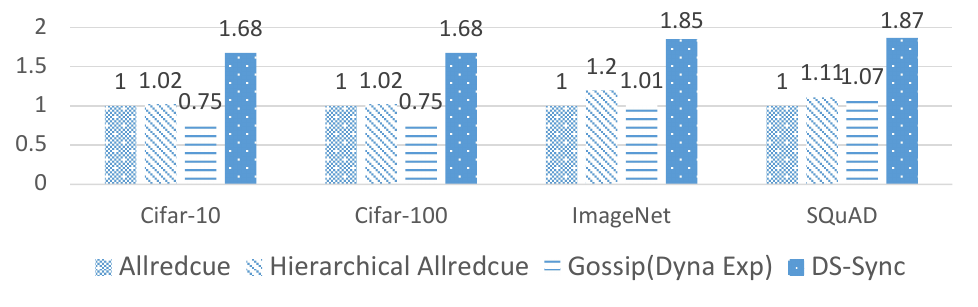}
	\caption{Speed up on TTA in inter-rack bandwidth contention}
	\label{fig:scenario_2}
\end{figure}
  DS-Sync keeps the bottlenecked inter-rack link in the group of 2 workers to alleviate the inter-rack network bottleneck. Fig.~\ref{fig:scenario_2} illustrates that DS-Sync also achieves the best end-to-end performance to reach the target accuracy in the scenario of inter-rack bandwidth contention.

 We evaluate all metrics on communication efficiency and convergence for all methods in this scenario, as shown in Tab.~\ref{tab:scenario_2}. The results verify that DS-Sync has the best communication efficiency by reducing inter-rack communication traffic and simultaneously synchronizing all inter-rack and intra-rack groups. Meanwhile, it also reaches the same accuracy as BSP methods within a similar iteration number.

In contrast, other baselines do not handle the bottleneck so well. The inter-rack bandwidth contention slows down the whole Allreduce pipeline. Although Hierarchical Allreduce has better communication efficiency by isolating the inter-rack bottleneck, it cannot simultaneously conduct inter-rack and intra-rack communication like our DS-Sync. ASP and Gossip can still worsen the inter-rack bottleneck. Once again, ASP fails to converge properly due to the high staleness.

\begin{table}[t]
\caption{Summary on communication efficiency and convergence with inter-rack bandwidth contention}
\label{tab:scenario_2}
\begin{tabular}{|c|cccc|}
\hline
Dataset                   & Methods  & Comm. Time(ms)       & Iter. \# & Best Acc.          \\ \hline
\multirow{5}{*}{Cifar10}  & AR       & 525.1                & 6560     & 92.51$\pm$0.92\%   \\
                          & Hier. AR & 515.3                & 6560     & 92.51$\pm$0.92\%   \\
                          & ASP(PS)  & \textgreater{}561.2  & -        & 88.17$\pm$1.58\%   \\
                          & Gossip   & 764.1                & 6360     & 92.47$\pm$1.17\%   \\
                          & DS-Sync  & \textbf{331.7}       & 5640     & 92.79$\pm$0.56\%   \\ \hline
\multirow{5}{*}{Cifar100} & AR       & 531.6                & 12740    & 76.79$\pm$0.28\%   \\
                          & Hier. AR & 523.5                & 12740    & 76.79$\pm$0.28\%   \\
                          & ASP(PS)  & \textgreater{}581.5  & -        & 68.19$\pm$2.51\%   \\
                          & Gossip   & 768.3                & 12130    & 76.66$\pm$0.32\%   \\
                          & DS-Sync  & \textbf{336.1}       & 11460    & 76.86$\pm$0.31\%   \\ \hline
\multirow{5}{*}{ImageNet} & AR       & 823.1                & 341,664  & 76.27\%            \\
                          & Hier. AR & 685.2                & 341,664  & 76.31\%            \\
                          & ASP(PS)  & \textgreater{}701.9  & -        & 73.50\%            \\
                          & Gossip   & 812.9                & 341,664  & 76.10\%            \\
                          & DS-Sync  & \textbf{407.4}       & 341,664  & 76.38\%            \\ \hline
\multirow{6}{*}{SQuAD}    & Methods  & Comm. Time(ms)       & Iter. \# & Best F1            \\ \cline{2-5} 
                          & AR       & 3129.3               & 7,400    & 92.66$\pm$0.08\%   \\
                          & Hier. AR & 2801.5               & 7,400    & 92.66$\pm$0.08\%   \\
                          & ASP(PS)  & \textgreater{}2783.1 & -        & 90.27\%$\pm$0.51\% \\
                          & Gossip   & 2913.1               & 7,400    & 92.61$\pm$0.1\%    \\
                          & DS-Sync  & \textbf{1633.4}      & 7,400    & 92.68$\pm$0.05\%   \\ \hline
\end{tabular}
\end{table}

\subsubsection{Static Topology Heterogeneity} 
\begin{figure}[t]
	\centering 
	\includegraphics[width=0.95\linewidth]{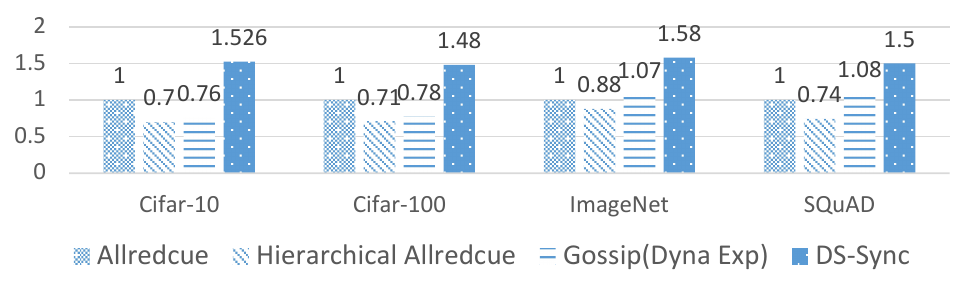}
	\caption{Speed up on TTA in static topology heterogeneity}
	\label{fig:scenario_3}
\end{figure}

In the group initialization, DS-Sync divides workers into one inter-rack group and two intra-rack groups to synchronize in parallel and avoid the oversubscription problem. Fig.~\ref{fig:scenario_3} shows that DS-Sync still reaches the target accuracy in the minimum time in the static hierarchical topology without any contention. 

 Tab.~\ref{tab:scenario_3} summarizes all metrics on communication efficiency and convergence for all methods for the static hierarchical topology. It shows that DS-Sync has advantages in communication efficiency since it synchronizes multiple smaller groups in parallel instead of the global one. DS-Sync also maintains the same convergence accuracy and rate as BSP methods.
 
 The baselines cannot simultaneously realize both aforementioned goals as DS-Sync does. Allreduce takes a longer time for global synchronization. Since Allreduce is free of oversubscriptions, Hierarchical Allreduce pays a higher communication cost than Allredcue for serial communications. ASP in PS architecture and Gossip in dynamic exponential graphs suffer from the oversubscription problem due to their multiple inter-rack connections. Additionally, the convergence of ASP is still affected by the oversubscription.
 \begin{table}[]
\caption{Summary on communication efficiency and convergence in static topology heterogeneity}
\label{tab:scenario_3}
\begin{tabular}{|c|cccc|}
\hline
Dataset                   & \multicolumn{1}{l}{Methods} & \multicolumn{1}{l}{Comm. Time(ms)} & \multicolumn{1}{l}{Iter. \#} & \multicolumn{1}{l|}{Best Acc.} \\ \hline
\multirow{5}{*}{Cifar10}  & AR                          & 266.1                              & 6560                         & 92.51$\pm$0.92\%               \\
                          & Hier. AR                    & 390.8                              & 6560                         & 92.51$\pm$0.92\%               \\
                          & ASP(PS)                     & \textgreater{}637.3                & -                            & 88.85$\pm$1.45\%               \\
                          & Gossip                      & 376.9                              & 6360                         & 92.47$\pm$1.17\%               \\
                          & DS-Sync                     & \textbf{200.1}                     & 5640                         & 92.79$\pm$0.56\%               \\ \hline
\multirow{5}{*}{Cifar100} & AR                          & 271.3                              & 12740                        & 76.79$\pm$0.28\%               \\
                          & Hier. AR                    & 395.1                              & 12740                        & 76.79$\pm$0.28\%               \\
                          & ASP(PS)                     & \textgreater{}641.2                & -                            & 69.41$\pm$1.81\%               \\
                          & Gossip                      & 374.8                              & 12130                        & 76.66$\pm$0.32\%               \\
                          & DS-Sync                     & \textbf{198.1}                     & 11460                        & 76.86$\pm$0.31\%               \\ \hline
\multirow{5}{*}{ImageNet} & AR                          & 441                                & 341,664                      & 76.27\%                        \\
                          & Hier. AR                    & 506.6                              & 341,664                      & 76.31\%                        \\
                          & ASP(PS)                     & \textgreater{}785.5                & -                            & 73.70\%                        \\
                          & Gossip                      & 405.3                              & 341,664                      & 76.10\%                        \\
                          & DS-Sync                     & \textbf{249.2}                     & 341,664                      & 76.38\%                        \\ \hline
\multirow{6}{*}{SQuAD}    & Methods                     & Comm. Time(ms)                     & Iter. \#                     & Best F1                        \\ \cline{2-5} 
                          & AR                          & 1631.8                             & 7,400                        & 92.66$\pm$0.08\%               \\
                          & Hier. AR                    & 2233.4                             & 7,400                        & 92.66$\pm$0.08\%               \\
                          & ASP(PS)                     & \textgreater{}3146.9               & -                            & 90.30\%$\pm$0.47\%             \\
                          & Gossip                      & 1502.6                             & 7,400                        & 92.61$\pm$0.1\%                \\
                          & DS-Sync                     & \textbf{1060.2}                    & 7,400                        & 92.68$\pm$0.05\%               \\ \hline
\end{tabular}
\end{table}

\section{Related Work}
There have been a large body of previous works related to DS-Sync, including but not limited to:
\begin{itemize}
\item \textbf{Hierarchical Allreduce:} The topology-aware Hierarchical Allreduce\cite{DBLP:conf/mlsys/ChoF0H19, DBLP:conf/mlsys/LuoWKCN20,DBLP:conf/apnet/WanZWHZ020} also reduces the communication traffic crossing the inter-rack link. Essentially, Hierarchical Allreduce still stalls or slows down other regular workers in serial or pipelined communication steps. In contrast, our DS-Sync divides workers into multiple non-overlapped groups to synchronize in parallel. 
\item \textbf{ASP and SSP:} ASP and SSP passively avoid or defer idle waiting for stochastic stragglers by relaxing the global barrier. However, network bottlenecks accumulate the staleness to a high level beyond their expectation, which violates their implicit assumption that stragglers happen on different nodes in different iterations. The accumulated and consistent staleness slows down the convergence speed and damages accuracy. Furthermore, high staleness also frequently exceeds SSP’s bound to trigger global synchronization just like BSP. However, our DS-Sync actively reduces the traffic on the network bottleneck with the divide-and-shuffle group pattern so that it improves communication efficiency without loss in convergence speed and accuracy.
\item \textbf{Gossip:} In Gossip, every worker only point-to-point communicates with neighbors in the sparsely connected graph of all workers. Gossip is not topology-aware so that it can have multiple connections crossing inter-rack links, especially for a time-varied graph. Point-to-point communication can be inefficient if some workers have multiple neighbors, which makes those be the bottleneck. But DS-Sync mitigates the network bottlenecks by keeping them in a smaller group and it also enjoys the benefits of allreduce communication within one group.
\item \textbf{Scheduling for Affinity:} Some works schedule the worker placement by considering the affinity in the hierarchical network~\cite{DBLP:conf/usenix/JeonVPQXY19,DBLP:conf/osdi/ZhaoHYZYZYLWXW20,DBLP:conf/osdi/XiaoBRSKHPPZZYZ18}. However, it is impossible to guarantee that all workers always can be in the same region and free of interference from other tasks. Our DS-Sync is complimentary when good affinity is impossible.  
\item \textbf{Reducing Communication:} Previous works also try to make use of gradient sparsification\cite{DBLP:conf/iclr/LinHM0D18}, quantization\cite{DBLP:conf/nips/AlistarhG0TV17}, and hierarchical representations in NN layers\cite{DBLP:journals/corr/abs-2008-08445,DBLP:conf/apnet/XiaZZW0J019} and local SGD\cite{DBLP:conf/iclr/Stich19} to reduce the communication. They discard, compress, or merge the trivial information and only transfer the necessary information to save communication costs. The line of works has the same communication cost for every worker and they are orthogonal to our DS-Sync.
\item \textbf{Underlying Network Optimization:} Some works conduct optimizations on underlying network, including overlapping communication with computation\cite{DBLP:conf/ppopp/AwanHHP17,DBLP:conf/usenix/ZhangZXDHLHWXX17,DBLP:conf/infocom/BaoPCW20,DBLP:conf/sosp/PengZCBYLWG19,DBLP:conf/apnet/XiaZZW0J019,DBLP:journals/corr/abs-2008-08445}, 
exploiting fast network protocols like RDMA\cite{DBLP:conf/sigcomm/GuoWDSYPL16,DBLP:conf/sigcomm/YiXC017},
performing in-network aggregation for less traffic\cite{DBLP:journals/corr/abs-1803-01491,DBLP:conf/nsdi/LaoLMCWAS21,DBLP:conf/nsdi/SapioC0NKKKMPR21}, 
minimizing network flow completion time by congestion control\cite{DBLP:conf/sigcomm/AlizadehGMPPPSS10,DBLP:conf/hotnets/ChenH0WT13,zeng2022cutting}, 
flow scheduling\cite{DBLP:conf/nsdi/Bai0WCHT15,DBLP:conf/infocom/LiBCHZLY17,DBLP:conf/sigcomm/ChenL0L18}, or coflow scheduling\cite{DBLP:conf/sigcomm/ChowdhuryZS14,DBLP:conf/sigcomm/ZhangCY0CG16,DBLP:conf/infocom/Zhao00YTGZLW15}. 
These underlying network optimizations can be further integrated with DS-Sync as a whole system.

\end{itemize}

\section{Conclusions}
We proposed the new DS-Sync to address network bottlenecks in the production cluster. DS-Sync divides and shuffles workers to form groups of different sizes. The groups are non-overlapping and synchronized in parallel. In theory, we quantitatively analyze its communication cost and mathematically prove its convergence. Furthermore, we conduct extensive testbed experiments to validate that DS-Sync can achieve communication efficiency and convergence accuracy simultaneously.

\bibliographystyle{unsrt}
\bibliography{references.bib}

\begin{thebibliography}{10}

\bibitem{DBLP:conf/usenix/JeonVPQXY19}
Myeongjae Jeon, Shivaram Venkataraman, Amar Phanishayee, Junjie Qian, Wencong
  Xiao, and Fan Yang.
\newblock Analysis of large-scale multi-tenant {GPU} clusters for {DNN}
  training workloads.
\newblock In {\em {ATC}}, 2019.

\bibitem{DBLP:conf/osdi/ZhaoHYZYZYLWXW20}
Hanyu Zhao, Zhenhua Han, Zhi Yang, Quanlu Zhang, Fan Yang, Lidong Zhou, Mao
  Yang, Francis C.~M. Lau, Yuqi Wang, Yifan Xiong, and Bin Wang.
\newblock Hived: Sharing a {GPU} cluster for deep learning with guarantees.
\newblock In {\em {OSDI}}, 2020.

\bibitem{DBLP:conf/mlsys/LuoWKCN20}
Liang Luo, Peter West, Arvind Krishnamurthy, Luis Ceze, and Jacob Nelson.
\newblock Plink: Discovering and exploiting locality for accelerated
  distributed training on the public cloud.
\newblock In {\em MLSys}, 2020.

\bibitem{DBLP:conf/osdi/XiaoBRSKHPPZZYZ18}
Wencong Xiao, Romil Bhardwaj, Ramachandran Ramjee, Muthian Sivathanu, Nipun
  Kwatra, Zhenhua Han, Pratyush Patel, Xuan Peng, Hanyu Zhao, Quanlu Zhang, Fan
  Yang, and Lidong Zhou.
\newblock Gandiva: Introspective cluster scheduling for deep learning.
\newblock In {\em {OSDI}}, 2018.

\bibitem{osdi:parameterserver}
Mu~Li, David~G. Andersen, Jun~Woo Park, Alexander~J. Smola, Amr Ahmed, Vanja
  Josifovski, James Long, Eugene~J. Shekita, and Bor{-}Yiing Su.
\newblock Scaling distributed machine learning with the parameter server.
\newblock In {\em OSDI}, 2014.

\bibitem{DBLP:conf/ppopp/AwanHHP17}
Ammar~Ahmad Awan, Khaled Hamidouche, Jahanzeb~Maqbool Hashmi, and
  Dhabaleswar~K. Panda.
\newblock S-caffe: Co-designing {MPI} runtimes and caffe for scalable deep
  learning on modern {GPU} clusters.
\newblock In {\em {SIGPLAN}}, 2017.

\bibitem{DBLP:conf/usenix/ZhangZXDHLHWXX17}
Hao Zhang, Zeyu Zheng, Shizhen Xu, Wei Dai, Qirong Ho, Xiaodan Liang, Zhiting
  Hu, Jinliang Wei, Pengtao Xie, and Eric~P. Xing.
\newblock Poseidon: An efficient communication architecture for distributed
  deep learning on {GPU} clusters.
\newblock In {\em {ATC}}, 2017.

\bibitem{DBLP:conf/infocom/BaoPCW20}
Yixin Bao, Yanghua Peng, Yangrui Chen, and Chuan Wu.
\newblock Preemptive all-reduce scheduling for expediting distributed {DNN}
  training.
\newblock In {\em {INFOCOM}}, 2020.

\bibitem{DBLP:conf/sosp/PengZCBYLWG19}
Yanghua Peng, Yibo Zhu, Yangrui Chen, Yixin Bao, Bairen Yi, Chang Lan, Chuan
  Wu, and Chuanxiong Guo.
\newblock A generic communication scheduler for distributed {DNN} training
  acceleration.
\newblock In {\em {SOSP}}, 2019.

\bibitem{DBLP:conf/cluster/MamidalaLP04}
Amith~R. Mamidala, Jiuxing Liu, and Dhabaleswar~K. Panda.
\newblock Efficient barrier and allreduce on infiniband clusters using
  multicast and adaptive algorithms.
\newblock In {\em {CLUSTER}}, 2004.

\bibitem{DBLP:conf/mlsys/ChoF0H19}
Minsik Cho, Ulrich Finkler, David~S. Kung, and Hillery~C. Hunter.
\newblock Blueconnect: Decomposing all-reduce for deep learning on
  heterogeneous network hierarchy.
\newblock In {\em MLSys}. mlsys.org, 2019.

\bibitem{asp}
Benjamin Recht, Christopher R{\'{e}}, Stephen~J. Wright, and Feng Niu.
\newblock Hogwild: {A} lock-free approach to parallelizing stochastic gradient
  descent.
\newblock In {\em NeurIPS}, 2011.

\bibitem{nips:ssp}
Qirong Ho, James Cipar, Henggang Cui, Seunghak Lee, Jin~Kyu Kim, Phillip~B.
  Gibbons, Garth~A. Gibson, Gregory~R. Ganger, and Eric~P. Xing.
\newblock More effective distributed {ML} via a stale synchronous parallel
  parameter server.
\newblock In {\em NeurIPS}, 2013.

\bibitem{DBLP:conf/ipps/PatarasukY07}
Pitch Patarasuk and Xin Yuan.
\newblock Bandwidth efficient all-reduce operation on tree topologies.
\newblock In {\em {IPDPS}}, 2007.

\bibitem{DBLP:conf/ecms/BilalKKZHMMWC12}
Kashif Bilal, Samee~Ullah Khan, Joanna Kolodziej, Limin Zhang, Khizar Hayat,
  Sajjad~Ahmad Madani, Nasro Min{-}Allah, Lizhe Wang, and Dan Chen.
\newblock A comparative study of data center network architectures.
\newblock In {\em ECMS}, 2012.

\bibitem{DBLP:conf/osdi/JiangZLYCG20}
Yimin Jiang, Yibo Zhu, Chang Lan, Bairen Yi, Yong Cui, and Chuanxiong Guo.
\newblock A unified architecture for accelerating distributed {DNN} training in
  heterogeneous {GPU/CPU} clusters.
\newblock In {\em {OSDI}}, 2020.

\bibitem{DBLP:conf/apnet/XiaZZW0J019}
Jiacheng Xia, Gaoxiong Zeng, Junxue Zhang, Weiyan Wang, Wei Bai, Junchen Jiang,
  and Kai Chen.
\newblock Rethinking transport layer design for distributed machine learning.
\newblock In {\em APNet}, pages 22--28. {ACM}, 2019.

\bibitem{DBLP:journals/corr/abs-2008-08445}
Hao Wang, Jingrong Chen, Xinchen Wan, Han Tian, Jiacheng Xia, Gaoxiong Zeng,
  Weiyan Wang, Kai Chen, Wei Bai, and Junchen Jiang.
\newblock Domain-specific communication optimization for distributed {DNN}
  training.
\newblock {\em CoRR}, abs/2008.08445, 2020.

\bibitem{DBLP:conf/ppopp/LiBGAH20}
Shigang Li, Tal Ben{-}Nun, Salvatore~Di Girolamo, Dan Alistarh, and Torsten
  Hoefler.
\newblock Taming unbalanced training workloads in deep learning with partial
  collective operations.
\newblock In {\em PPoPP}, 2020.

\bibitem{DBLP:conf/nips/LianZZHZL17}
Xiangru Lian, Ce~Zhang, Huan Zhang, Cho{-}Jui Hsieh, Wei Zhang, and Ji~Liu.
\newblock Can decentralized algorithms outperform centralized algorithms? {A}
  case study for decentralized parallel stochastic gradient descent.
\newblock In {\em NeurIPS}, 2017.

\bibitem{DBLP:conf/icml/AssranLBR19}
Mahmoud Assran, Nicolas Loizou, Nicolas Ballas, and Michael~G. Rabbat.
\newblock Stochastic gradient push for distributed deep learning.
\newblock In {\em ICML}, 2019.

\bibitem{DBLP:conf/sigcomm/GuoWDSYPL16}
Chuanxiong Guo, Haitao Wu, Zhong Deng, Gaurav Soni, Jianxi Ye, Jitu Padhye, and
  Marina Lipshteyn.
\newblock {RDMA} over commodity ethernet at scale.
\newblock In {\em SIGCOMM}, pages 202--215. {ACM}, 2016.

\bibitem{DBLP:conf/sigcomm/YiXC017}
Bairen Yi, Jiacheng Xia, Li~Chen, and Kai Chen.
\newblock Towards zero copy dataflows using {RDMA}.
\newblock In {\em SIGCOMM Posters and Demos}. {ACM}, 2017.

\bibitem{DBLP:journals/corr/abs-1803-01491}
Li~Chen, Ge~Chen, Justinas Lingys, and Kai Chen.
\newblock Programmable switch as a parallel computing device.
\newblock {\em CoRR}, abs/1803.01491, 2018.

\bibitem{DBLP:conf/nsdi/LaoLMCWAS21}
ChonLam Lao, Yanfang Le, Kshiteej Mahajan, Yixi Chen, Wenfei Wu, Aditya Akella,
  and Michael~M. Swift.
\newblock {ATP:} in-network aggregation for multi-tenant learning.
\newblock In {\em NSDI}. {USENIX} Association, 2021.

\bibitem{DBLP:conf/nsdi/SapioC0NKKKMPR21}
Amedeo Sapio, Marco Canini, Chen{-}Yu Ho, Jacob Nelson, Panos Kalnis, Changhoon
  Kim, Arvind Krishnamurthy, Masoud Moshref, Dan R.~K. Ports, and Peter
  Richt{\'{a}}rik.
\newblock Scaling distributed machine learning with in-network aggregation.
\newblock In {\em NSDI}. {USENIX} Association, 2021.

\bibitem{DBLP:conf/sigcomm/AlizadehGMPPPSS10}
Mohammad Alizadeh, Albert~G. Greenberg, David~A. Maltz, Jitendra Padhye,
  Parveen Patel, Balaji Prabhakar, Sudipta Sengupta, and Murari Sridharan.
\newblock Data center {TCP} {(DCTCP)}.
\newblock In {\em SIGCOMM}. {ACM}, 2010.

\bibitem{DBLP:conf/hotnets/ChenH0WT13}
Li~Chen, Shuihai Hu, Kai Chen, Haitao Wu, and Danny H.~K. Tsang.
\newblock Towards minimal-delay deadline-driven data center {TCP}.
\newblock In {\em HotNets}, pages 21:1--21:7. {ACM}, 2013.

\bibitem{DBLP:conf/nsdi/Bai0WCHT15}
Wei Bai, Kai Chen, Hao Wang, Li~Chen, Dongsu Han, and Chen Tian.
\newblock Information-agnostic flow scheduling for commodity data centers.
\newblock In {\em NSDI}. {USENIX} Association, 2015.

\bibitem{DBLP:conf/infocom/LiBCHZLY17}
Ziyang Li, Wei Bai, Kai Chen, Dongsu Han, Yiming Zhang, Dongsheng Li, and
  Hongfang Yu.
\newblock Rate-aware flow scheduling for commodity data center networks.
\newblock In {\em INFOCOM}. {IEEE}, 2017.

\bibitem{DBLP:conf/sigcomm/ChenL0L18}
Li~Chen, Justinas Lingys, Kai Chen, and Feng Liu.
\newblock Auto: scaling deep reinforcement learning for datacenter-scale
  automatic traffic optimization.
\newblock In {\em SIGCOMM}. {ACM}, 2018.

\bibitem{DBLP:conf/sigcomm/ChowdhuryZS14}
Mosharaf Chowdhury, Yuan Zhong, and Ion Stoica.
\newblock Efficient coflow scheduling with varys.
\newblock In {\em SIGCOMM}. {ACM}, 2014.

\bibitem{DBLP:conf/sigcomm/ZhangCY0CG16}
Hong Zhang, Li~Chen, Bairen Yi, Kai Chen, Mosharaf Chowdhury, and Yanhui Geng.
\newblock {CODA:} toward automatically identifying and scheduling coflows in
  the dark.
\newblock In {\em SIGCOMM}. {ACM}, 2016.

\bibitem{DBLP:conf/infocom/Zhao00YTGZLW15}
Yangming Zhao, Kai Chen, Wei Bai, Minlan Yu, Chen Tian, Yanhui Geng, Yiming
  Zhang, Dan Li, and Sheng Wang.
\newblock Rapier: Integrating routing and scheduling for coflow-aware data
  center networks.
\newblock In {\em INFOCOM}. {IEEE}, 2015.

\bibitem{DBLP:conf/apnet/WanZWHZ020}
Xinchen Wan, Hong Zhang, Hao Wang, Shuihai Hu, Junxue Zhang, and Kai Chen.
\newblock {RAT} - resilient allreduce tree for distributed machine learning.
\newblock In {\em APNet}. {ACM}, 2020.

\bibitem{DBLP:conf/imw/SarvothamRB01}
Shriram Sarvotham, Rudolf~H. Riedi, and Richard~G. Baraniuk.
\newblock Connection-level analysis and modeling of network traffic.
\newblock In {\em SIGCOMM}, 2001.

\bibitem{DBLP:conf/iclr/DaiZDZX19}
Wei Dai, Yi~Zhou, Nanqing Dong, Hao Zhang, and Eric~P. Xing.
\newblock Toward understanding the impact of staleness in distributed machine
  learning.
\newblock In {\em ICLR}, 2019.

\bibitem{wang2019matcha}
Jianyu Wang, Anit~Kumar Sahu, Zhouyi Yang, Gauri Joshi, and Soummya Kar.
\newblock Matcha: Speeding up decentralized sgd via matching decomposition
  sampling.
\newblock In {\em ICC}. IEEE, 2019.

\bibitem{DBLP:conf/uai/IzmailovPGVW18}
Pavel Izmailov, Dmitrii Podoprikhin, Timur Garipov, Dmitry~P. Vetrov, and
  Andrew~Gordon Wilson.
\newblock Averaging weights leads to wider optima and better generalization.
\newblock In {\em {UAI}}, 2018.

\bibitem{DBLP:conf/iclr/KeskarMNST17}
Nitish~Shirish Keskar, Dheevatsa Mudigere, Jorge Nocedal, Mikhail Smelyanskiy,
  and Ping Tak~Peter Tang.
\newblock On large-batch training for deep learning: Generalization gap and
  sharp minima.
\newblock In {\em ICLR}, 2017.

\bibitem{DBLP:journals/siamrev/BottouCN18}
L{\'{e}}on Bottou, Frank~E. Curtis, and Jorge Nocedal.
\newblock Optimization methods for large-scale machine learning.
\newblock {\em {SIAM} Rev.}, 2018.

\bibitem{DBLP:journals/pieee/NedicOR18}
Angelia Nedic, Alex Olshevsky, and Michael~G. Rabbat.
\newblock Network topology and communication-computation tradeoffs in
  decentralized optimization.
\newblock {\em Proc. {IEEE}}, 2018.

\bibitem{DBLP:conf/bmvc/ZagoruykoK16}
Sergey Zagoruyko and Nikos Komodakis.
\newblock Wide residual networks.
\newblock In {\em {BMVC}}, 2016.

\bibitem{cifardataset}
Alex Krizhevsky, Vinod Nair, and Geoffrey Hinton.
\newblock The cifar-10 dataset.
\newblock {\em online: http://www. cs. toronto. edu/kriz/cifar. html}, 55,
  2014.

\bibitem{DBLP:conf/cvpr/DengDSLL009}
Jia Deng, Wei Dong, Richard Socher, Li{-}Jia Li, Kai Li, and Fei{-}Fei Li.
\newblock Imagenet: {A} large-scale hierarchical image database.
\newblock In {\em {(CVPR} 2009)}.

\bibitem{DBLP:conf/naacl/DevlinCLT19}
Jacob Devlin, Ming{-}Wei Chang, Kenton Lee, and Kristina Toutanova.
\newblock {BERT:} pre-training of deep bidirectional transformers for language
  understanding.
\newblock In {\em {NAACL-HLT}}, 2019.

\bibitem{DBLP:conf/emnlp/RajpurkarZLL16}
Pranav Rajpurkar, Jian Zhang, Konstantin Lopyrev, and Percy Liang.
\newblock Squad: 100, 000+ questions for machine comprehension of text.
\newblock In {\em {EMNLP}}, 2016.

\bibitem{DBLP:conf/iclr/LinHM0D18}
Yujun Lin, Song Han, Huizi Mao, Yu~Wang, and Bill Dally.
\newblock Deep gradient compression: Reducing the communication bandwidth for
  distributed training.
\newblock In {\em ICLR}. OpenReview.net, 2018.

\bibitem{DBLP:conf/nips/AlistarhG0TV17}
Dan Alistarh, Demjan Grubic, Jerry Li, Ryota Tomioka, and Milan Vojnovic.
\newblock {QSGD:} communication-efficient {SGD} via gradient quantization and
  encoding.
\newblock In {\em NeurIPS}, 2017.

\bibitem{DBLP:conf/iclr/Stich19}
Sebastian~U. Stich.
\newblock Local {SGD} converges fast and communicates little.
\newblock In {\em ICLR}. OpenReview.net, 2019.

\bibitem{zeng2022cutting}
Gaoxiong Zeng, Li~Chen, Bairen Yi, and Kai Chen.
\newblock Cutting tail latency in commodity datacenters with cloudburst.
\newblock In {\em IEEE INFOCOM}, 2022.

\end{thebibliography}

\end{document}